# Development of a hybrid learning system based on SVM, ANFIS and domain knowledge: DKFIS

Soumi Chaki[1], Aurobinda Routray[1], William K. Mohanty[2], Mamata Jenamani[3]

[1]Department of Electrical Engineering,
IIT Kharagpur, India
soumibesu2008@gmail.com
aroutray@ee.iitkgp.ernet.in

[2] Department of Geology and Geophysics,
IIT Kharagpur, India
wkmohanty@gg.iitkgp.ernet.in

[3]Department of Industrial and Systems Engineering,
IIT Kharagpur, India
mj@iem.iitkgp.ernet.in

*Abstract*— This paper presents the development of a hybrid learning system based on Support Vector Machines (SVM), Adaptive Neuro-Fuzzy Inference System (ANFIS) and domain knowledge to solve prediction problem. The proposed two-stage Domain Knowledge based Fuzzy Information System (DKFIS) improves the prediction accuracy attained by ANFIS alone. The proposed framework has been implemented on a noisy and incomplete dataset acquired from a hydrocarbon field located at western part of India. Here, oil saturation has been predicted from four different well logs i.e. gamma ray, resistivity, density, and clay volume. In the first stage, depending on zero or near zero and non-zero oil saturation levels the input vector is classified into two classes (Class 0 and Class 1) using SVM. The classification results have been further fine-tuned applying expert knowledge based on the relationship among predictor variables i.e. well logs and target variable - oil saturation. Second, an ANFIS is designed to predict non-zero (Class 1) oil saturation values from predictor logs. The predicted output has been further refined based on expert knowledge. It is apparent from the experimental results that the expert intervention with qualitative judgment at each stage has rendered the prediction into the feasible and realistic ranges. The performance analysis of the prediction in terms of four performance metrics such as correlation coefficient (CC), root mean square error (RMSE), and absolute error mean (AEM), scatter index (SI) has established DKFIS as a useful tool for reservoir characterization.

*Keywords—Support Vector Machine (SVM), Adaptive Neuro-fuzzy Inference System (ANFIS), knowledge base, Domain Knowledge based Fuzzy Information System (DKFIS), qualitative learning, g-metric means, performance metrics.*

I. INTRODUCTION

The design of an appropriate framework includes the selection of appropriate pre-processing and learning algorithms. On the other hand, selection of relevant algorithms requires information on the dataset and domain knowledge related to the problem. Therefore, devising an efficient framework is a two-way process. First, the algorithms are selected based on dataset characteristics, literature review, and scope to incorporate machine learning algorithms to address the problem. Then, the parameters associated with the learning algorithms can be tuned depending upon the performance obtained from the designed framework. Thus, the framework is finalized iteratively and applied.

This study reveals design steps of an advanced prediction framework namely DKFIS which is a fusion of SVM, ANFIS, and domain knowledge. Then, the designed framework is implemented on a noisy and incomplete geological dataset to solve a real-world problem i.e. prediction of oil saturation from multiple well logs. Therefore, knowledge on the reservoir characterization domain such as inherent relationship between lithological characteristics and well logs has been utilized to assist the prediction algorithms.

Predicting reservoir properties from well logs [1]–[4] and seismic attributes [5]–[7] is a well-known problem. The main target properties include porosity, permeability, oil saturation etc. It has been found in existing literatures that gamma ray, resistivity, density, and sonic logs have been used as predictor variables in [8]–[10]. However, the choice of predictor variables also depends on availability of the dataset.

The solution approaches for the prediction problem under consideration include use of various soft-computing tools such as Artificial Neural Network (ANN), Fuzzy Logic and Genetic Algorithms (GA) and several hybrid algorithms. For example, to predict permeability Improved Fuzzy Neural Network (IFNN) based on a gradient descent Fuzzy algorithm [11], and GA with Neuro-Fuzzy techniques [12], [13] have been used in the past. The performance of a particular technique depends on nature of the dataset. Though, attempts have been made to use novel machine learning techniques in search of better solution, sometimes relatively older techniques produce better or comparable solution with reduced computational complexity compared to their recent counterparts. For example, both ANN and ANFIS has been used to estimate lithological properties from well logs in [14]–[16] yielding acceptable results. Apart from the commonly used approaches, sometimes relatively uncommon methods are also selected to predict lithological properties. For example, the oil saturation has been estimated from carbon/ oxygen logs using Monte Carlo method in [17].

The literature reports considerable research in the applications of a wide variety of methods starting from statistical regression to soft computing for modeling various parameters related to reservoir. Neural Network, Fuzzy Logic and Neuro-Fuzzy methods have been very popular in reservoir characterization in presence of uncertain and incomplete data. These methods predominantly use statistical and data intensive machine learning tools. Typically, these algorithms suffer from the curse of dimensionality. There are still lot of gap areas which needs immediate attention. In this paper, we have taken up the following research issues.

a. Learning in presence of sparse and unevenly distributed data
b. Inclusion of both quantitative and qualitative domain knowledge

The well log data are generated from a complex, heterogeneous and non-linear system. Sometimes, measurement error and noise deteriorate the quality of the dataset. The choice of machine learning method and thereby prediction accuracy is greatly dependent on the data-set. In most cases preprocessing is carried out before using the prediction model. The selection of proper pre-processing algorithms improves the performance of machine learning algorithms to a great extent. Therefore, it is essential to analyze the nature of the working dataset and carry out suitable pre-processing before entering into prediction stage. For example, regularization scheme in pre-processing stage has improved the prediction capability of ANN as in [18]. Similarly, normalization, relevant attribute selections have been carried out to improve the performance of the classification framework in [9]. In another domain [19], [20], band pass filtering has been used to remove artifacts from raw EEG data followed by normalization in pre-processing stage. The selection of appropriate parameters of the pre-processing techniques are dependent on the user. Therefore, knowledge on the dataset characteristics and the problem are essential to select the pre-processing algorithms and associated parameters wisely; which in turn is vital for performance of the learning framework. This realization has inspired the inclusion of expert knowledge explicitly in the process while designing the framework in this study.

Often the models generated using data-driven methods produce infeasible results when the training set is not dense enough. Under such circumstances the model turns out to be ill-conditioned and may yield unacceptable and random results. Expert judgment is necessary to make qualitative assessment of the fidelity of the result and modify them if necessary. Methods are available to incorporate domain knowledge into models generated primarily from data-driven methods. A concept of 'Qfilter' to achieve 'qualitatively faithful numerical prediction' is suggested by Suc et al. [21], [22].

In this paper, we propose a new approach to estimate oil saturation using four predictor variables i.e. gamma ray, resistivity, density, and clay fraction. The dataset under the study area is highly skewed in nature in the sense that around 93.55% of the patterns carry zero oil saturation level, which is quite common in fields producing depleting quantity of hydrocarbon. This dataset is observed to be imprecise and unreliable in the sense that almost all patterns correspond to a particular well in this dataset carry zero oil saturation level only, which makes it difficult to solely depend on the result of prediction model. To circumvent to this problem we use a two stage model. In the 1st stage, a SVM based classifier is trained to group the inputs that corresponds to the zero and non-zero oil saturation values. In the 2nd stage, a continuous Neuro-Fuzzy predictor which is used to precisely estimate only non-zero oil-saturation. We have used domain knowledge after each stage to fine tune the prediction results for making them more realistic.

Our approach is mostly suitable for sparse and unevenly distributed datasets where the output is clustered around few classes. In this case the output, oil saturation values are predominantly zero. Thus there is a very small representative training sets from the non-zero class to build the prediction model. Under these circumstances the domain knowledge is helpful in correcting the predictive models. The SVM is used to separate the patterns into two classes zero (Class 0) and non-zero (Class 1). The patterns corresponding to non-zero oil saturation values are used to train an ANFIS. A Fuzzy Inference System (FIS) is designed using the domain knowledge (say it Domain Knowledge based FIS (DKFIS)). DKFIS is applied to the outputs of both the stages to validate or suitably modify them. The dataset consists of well-logs such as gamma ray, resistivity, density, clay fraction and the corresponding oil-saturation.

The contribution of the paper can be outlined as follows:

a) Design of a two stage machine learning system for classifying and then predicting the oil saturation values
b) The use of DKFIS to improve the prediction results

The rest of the paper is structured as follows: we first give a brief description of the ANFIS; next an overview of classification based on SVM is provided; a brief discussion on the concept of qualitative learning and Qfilter is given in the section following. Then, we describe the dataset and knowledge base used in this study. Next, the proposed two-step methodology to model oil saturation from input logs and lithology log is described. In the following section, experimental results obtained by the proposed algorithm on current dataset is reported. Finally, the paper concludes with future scope of the current study.

II. QUALITATIVE LEARNING

The prediction accuracy of numerical learning models should be consistent with qualitative domain knowledge. A concept of numerical regression method namely 'Qfilter' as in [21], [22] is used here to integrate domain knowledge with prediction accuracy. A set of qualitative constraints derived from domain knowledge of a specific application is provided as input to 'Qfilter' along with numerical patterns. In contrast, 'Qfilter' can be applied within 'Q2' learning. For the latter case, no expert given qualitative constraints is required. In case of 'Qfilter', a qualitative tree is described which represents the relation between input attributes and class using Qualitatively Constrained Functions (QCF). QCF is said to be consistent if it is consistent with all possible example pairs. It can be said that 'Qfilter' is a filter which filters the qualitative errors caused by measurement errors. 'Qfilter' obtains improved accuracy consistently over standard regression models such as Locally Weighted Regression (LWR) irrespective of number of available patterns and noise level in the dataset. In this paper, the Qfilter is implemented as an FIS; thus the complete DKFIS framework is designed.

## III. DESCRIPTION OF THE DATASET

In this study, well log data from four closely spaced wells placed in Indian onshore are used; henceforth these wells to be referred as A, B, C, and D respectively. The input logs used in this study are gamma ray, resistivity, density, and clay volume. For this particular dataset, the target variable i.e. oil saturation value has a range of 0 to 0.86 with mean and variance of 0.0391 and 0.0124 per unit respectively. Histogram plot of oil saturation is presented in Fig. 1. It can be observed that a large number of data points corresponds to zero oil saturation value. Therefore, the classification of dataset depending on zero (Class 0) and non-zero oil saturation level (Class 1) is desirable. Table I represents the expert knowledge base provided for this study analyzing the working dataset. Table I represents the expert knowledge base provided for this study analyzing the working dataset.

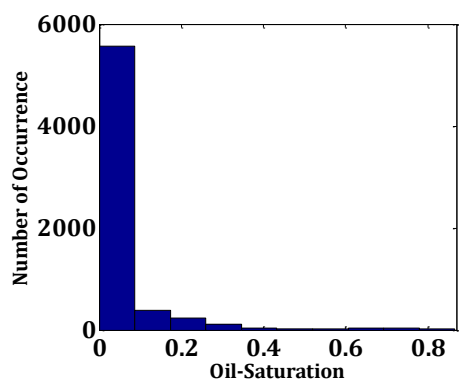

Fig. 1. The range distribution of oil saturation

TABLE I. EXPERT KNOWLEDGE BASE

| |
| --- |
| R1 : if <gamma ray is high, resistivity is low, density is high, and clay volume is high> then <oil saturation is low> |
| R2 : if < gamma ray is high, resistivity is high, density is high, and clay volume is low> then <oil saturation is low> |
| R3 : if < gamma ray is low, resistivity is medium, density is low, and clay volume is low> then <oil saturation is high> |

## IV. METHODOLOGY

First, the predictor variables e.g. gamma ray, resistivity etc. are normalized using z-score normalization as:

$$GR_{normalized} = \frac{GR - GR_{mean}}{GR_{standard\ deviation}} \quad (1)$$

In contrast, the target variable is normalized in [0 1] range using min-max normalization which essentially performs a linear transformation. The relationships among the original data values are preserved in this normalization. For an instance, $min_x$ and $max_x$ are minimum and maximum values of attribute $X$. This data interval $[min_x, max_x]$ is to be mapped into a new interval $[new\_min_x, new\_max_x]$. Therefore, every value $val$ from original data interval is mapped into value $normalized\_val$ using the formula:

$$normalized\_val = \frac{val - min_x}{max_x - min_x} * (new\_max_x - new\_min_x) + new\_min_x \quad (2)$$

As mentioned earlier for step-1, a classification model based on SVM has been constructed from the normalized data. The classification result of this stage has been further fine-tuned by DKFIS as will be discussed later. In the step-2, a non-linear relationship between predictor variables and target variable has been calibrated using Adaptive Neuro-Fuzzy Inference System (ANFIS). In order to train the ANFIS model, first, the normalized input variables have been fuzzified using generalized bell shaped membership functions (gbellmf) as:

$$f(x;a,b,c) = \frac{1}{1 + \left|\frac{x-c}{a}\right|^{2b}} \quad (3)$$

where '$c$' represents the center of the corresponding membership function, '$a$' represents half of the width; '$a$' and '$b$' together determine the slopes at the crossover points. The bell function is continuous and infinitely differentiable, which is required for gradient based learning. Fig. 2 represents bell shaped membership function where y-axis represents membership value of the fuzzified parameter at the corresponding data points and x axis represents universe of discourse. The parameters can be adjusted to tune the shape of the membership function.

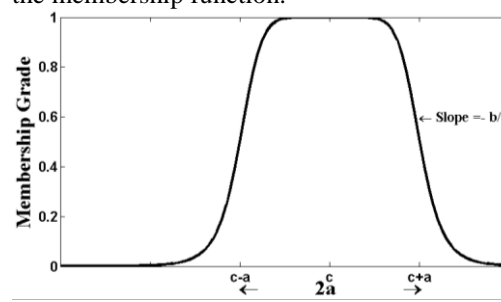

Fig. 2. Bell shaped membership function.

The ANFIS has been initialized and a hybrid algorithm combining least-squares and gradient descent method with back-propagation has been applied to train the parameters of FIS using a training dataset. The training patterns are the oil saturation values and associated inputs from Class-1.

### DKFIS FRAMEWORK

The qualitative domain knowledge as in Table I is coded into a FIS by fuzzyfying the oil saturation values. Fig. 3 represents the proposed hybrid SVM and ANFIS based cascaded learning system i.e. the DKFIS framework. Two membership functions have been defined depending on the range of oil saturation value- NZS (Non-zero Small) and NZB (Non-zero Big). Fig. 4 describes the plot of these two membership functions along the universe of discourse.

The classifier and the prediction models have been trained and tested using the dataset. The training set has been created by combining 70% patterns from each of the wells. The remaining 30% patterns from each of the wells have been combined to create the testing set. The predictor and target variables have been normalized using Z-score and min-max normalization respectively.

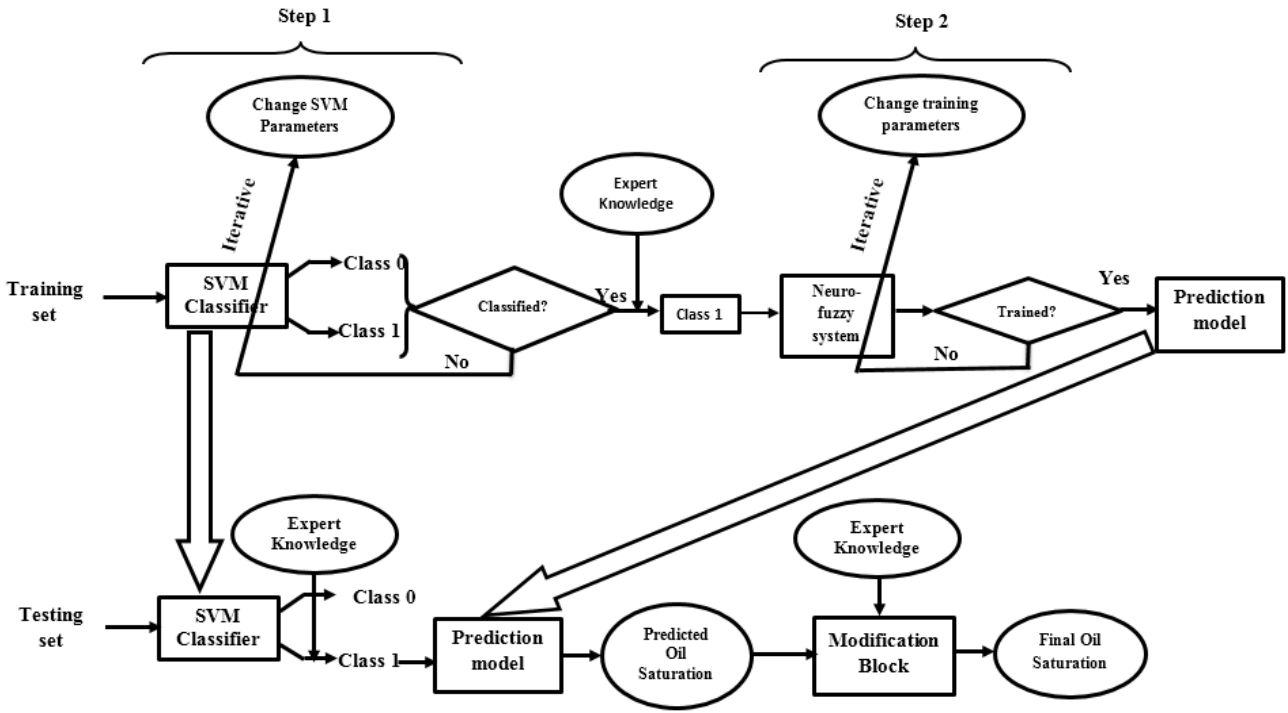

Fig. 3. Proposed hybrid SVM, ANFIS and domain knowledge based cascaded learning system (DKFIS)

For the particular dataset used in this study, the range of oil saturation value is from 0 to 0.86 with mean and variance of 0.0391 and 0.0124 per unit respectively. The membership functions NZS and NZB have been defined accordingly. As shown in Fig. 3, the outputs after Stage-1 as well as Stage-2 have been modified with the domain knowledge using the DKFIS in the following manner. First, the de-normalized prediction output has been fuzzified using the defied Gaussian membership functions NZB and NZS as in Fig. 4. For example, a pattern belongs to NZS category if fuzzified value of oil saturation by NZS membership function is more than the fuzzified value corresponding to NZB. Thus, the Class 1 patterns have been again classified into two categories- NZS (Non Zero Small) and NZB (Non Zero Big). Subsequently the membership grade of NZS and NZB have been modified using the domain knowledge from Table I. For example, Rule-1, and Rule-2 correspond to low oil saturation values for a given input condition. Suppose with these input conditions, the SVM categorizes the pattern as a Class 1 (which corresponds to high saturation values) sample instead of Class 0, then clearly the class label should be modified to Class 0 and the inputs should not be passed to the subsequent steps for prediction. These rules have been applied to each and every outcome and the membership grades have been changed as required. The modified oil saturation value has been obtained by defuzzifying the modified membership grades.

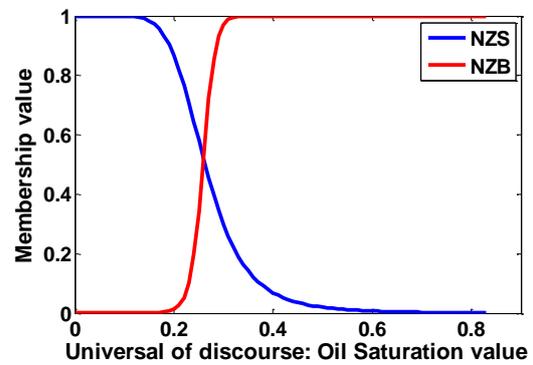

Fig. 4. Plot of membership functions depending on oil saturation value.

## V. EXPERIMENTAL RESULTS

First, the training and testing sets have been prepared as shown in Fig. 3 following the manner described in the previous section. Then, a SVM classifier has been calibrated to classify the patterns into Class 0 and Class 1 representing zero and non-zero oil saturation levels respectively using the training set. A fast algorithm namely Sequential Minimal Optimization (SMO) has been used to train the SVM model [23]. Then, the classification result has been modified by the expert knowledge as reported in Table I; followed by initialization and training of an ANFIS model using the training patterns belong to Class-1 resulted in step 1. Here, the well logs and oil saturation values have been used as predictors and target respectively. After completion of successful training, testing has been carried out. First, testing patterns have beem classified into Class 0 and Class 1 by calibrated SVM classifier followed by modification as suggested by expert knowledge. Then, the resulting Class 1 patterns have been evaluated by the calibrated ANFIS parameters. Finally, the prediction results have also been modified based on experts' knowledge. The improvement in the prediction has been quantified in terms of four metrics such as CC, RMSE, AEM and SI.

The classification performance of SVM classifier alone and including experts' knowledge have been reported in Table II in terms of g-metric means using the testing dataset. G-metric means is an appropriate parameter to quantify classification result while working with an imbalanced dataset [24]. The first three rows of Table II represent the performances of the same while working with different kernel functions e.g. Gaussian radial basis (rbf), linear, multilayer perceptron (mlp) respectively [19]. Gaussian rbf kernel outperforms others obtaining maximum g-metric means. The width of the kernel parameter has been also crucial to obtain high g-metric means. Fig. 5 represents the variation between g-metric means and rbf kernel parameter. It can be seen from Fig. 5 that the g-metric means is maximum while training the classifier with Gaussian rbf kernel with width parameter one. In each of the three cases, the g-metric means increases while including expert knowledge after the SVM.

Then, the trained ANFIS parameters have been used to obtain oil saturation values from the predictor variables of the testing patterns belong to Class 1 category as an outcome of step 1 in testing. Even though there seems to be a good testing performance in terms of the four metrics by ANFIS alone without including expert knowledge base as shown in Table III, the DKFIS has been used to integrate the domain knowledge for realistic predictions.

TABLE II. THE CLASSIFICATION PERFORMANCE OF SVM USING DIFFERENT KERNEL FUNCTION WITH THE TESTING PATTERNS

| Expert Knowledge | Kernel Function | G-metric means |
|---|---|---|
| Not Included | *rbf* | 0.9316 |
| | *linear* | 0.9197 |
| | *mlp* | 0.6293 |
| Included | *rbf* | 0.9397 |
| | *linear* | 0.9277 |
| | *mlp* | 0.6383 |

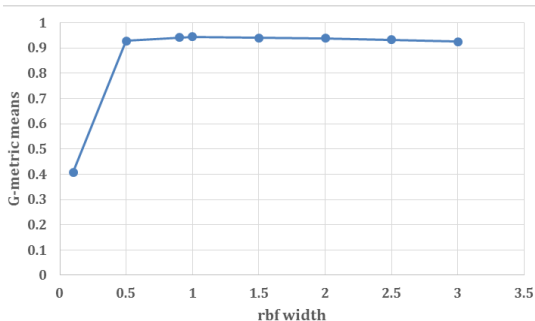

Fig. 5. Plot of G-metric means vs. rbf kernel parameter

TABLE III. STATISTICS OF TESTING PERFORMANCE IN PREDICTION

| Performance Indicators | Excluding Expert Knowledge | Including Expert Knowledge |
|---|---|---|
| CC | 0.91 | 0.95 |
| RMSE | 0.11 | 0.06 |
| AEM | 0.08 | 0.05 |
| SI | 0.48 | 0.21 |

These modifications using the DKFIS have been carried out by fuzzifying the predicted output using the Gaussian membership functions NZS (Non-zero Small) and NZB (Non-zero Big). In the cases where there is a mismatch in the prediction the membership grades of the predicted output have been modified. Finally, the modified prediction output has been achieved by defuzzifying the modified membership grades. The performance indicator values reported in Table III confirms the improvement in prediction while including expert opinions into consideration. Thus, the proposed DKFIS can be used as an efficient framework to predict lithological properties from well logs.

## VI. CONCLUSIONS

The proposed two-stage methodology can be applied to estimate lithological properties from input logs and lithology log for incomplete and imprecise dataset. In case of a good quality dataset also, the knowledge base learning would improve the prediction accuracy. In future, an initiative can be taken to design a prediction model to estimate petrophysical properties from seismic attributes using an expert defined knowledge base.

The proposed DKFIS has been implemented to solve a reservoir characterization problem efficiently. Therefore, it can be expected to carryout prediction effectively while working with dataset from other research areas. In those cases, the associated knowledge base would be developed by a domain knowledge expert.